\documentclass{article}

\usepackage[final]{neurips_2018}

\usepackage[utf8]{inputenc} 
\usepackage[T1]{fontenc}    
\usepackage{hyperref}       
\usepackage{url}            
\usepackage{booktabs}       
\usepackage{amsfonts}       
\usepackage{nicefrac}       
\usepackage{microtype}      
\usepackage{graphicx}
\usepackage{amsmath,amssymb}
\usepackage{amsfonts}       
\usepackage{color}
\usepackage{transparent}
\usepackage{natbib}

\usepackage{amsmath}
\usepackage{shortbold}
\usepackage{enumerate}

\usepackage{nicefrac}       

\usepackage{soul}

\usepackage{placeins}

\usepackage{subcaption}

\title{Moonshine: Distilling with Cheap Convolutions}

\author{
  Elliot J.~Crowley\\
  School of Informatics\\
  University of Edinburgh\\
  \texttt{elliot.j.crowley@ed.ac.uk} \\
  \And
  Gavin Gray\\
  School of Informatics\\
  University of Edinburgh\\
  \texttt{g.d.b.gray@ed.ac.uk} \\
  \And
   Amos Storkey\\
  School of Informatics\\
  University of Edinburgh\\
  \texttt{a.storkey@ed.ac.uk} \\
}

\begin{document}

\maketitle

\begin{abstract}
Many engineers wish to deploy modern neural networks in memory-limited
settings; but the development of flexible methods for reducing memory use is in
its infancy, and there is little knowledge of the resulting cost-benefit. We
propose structural model distillation for memory reduction using a strategy
that produces a \emph{student} architecture that is a simple transformation of
the \emph{teacher} architecture: no redesign is needed, and the same
hyperparameters can be used. Using attention transfer, we provide Pareto
curves/tables for distillation of residual networks with four benchmark
datasets, indicating the memory versus accuracy payoff. We show that
substantial memory savings are possible with very little loss of accuracy, and
confirm that distillation provides student network performance that is better
than training that student architecture directly on data.
\end{abstract}

\section{Introduction}

Despite advances in deep learning for a variety of tasks~\citep{lecun2015deep},
deployment of deep learning into embedded devices e.g.\ wearable devices,
digital cameras, vehicle navigation systems, has been relatively slow due to
resource constraints under which these devices operate. Big, memory-intensive
neural networks do not fit on these devices, but do these networks have to be
big and expensive? The dominant run-time memory cost of neural networks is the
number of parameters that need to be stored. Can we have networks with
substantially fewer parameters, without the commensurate loss of performance?

It is possible to take a large pre-trained \emph{teacher} network, and use its
outputs to aid in the training of a smaller \emph{student}
network~\citep{ba2013do} through some distillation process. By doing this the
student network is more powerful than if it was trained solely on the training
data, and is closer in performance to the larger teacher. The lower-parameter
student network typically has an architecture that is more shallow, or thinner
--- by which we mean its filters have fewer channels~\citep{romero2014fitnets}
--- than the teacher. While it is not possible to arbitrarily approximate any
network with another~\citep{urban2016do}, the limit in neural network
performance is at least in part due to the training algorithm, rather than its
representational power.

In this paper, we take an alternative approach in designing our student
networks. Instead of making networks thinner, or more shallow, we take the
standard convolutional block such networks possess and replace it with a
\emph{cheaper} convolutional block, keeping the original architecture. For
example, in a Residual Network (ResNet)~\citep{he2016deep} this standard block
is a pair of sequential 3$\times$3 convolutions. We show that for a comparable
number of parameters, student networks that retain the architecture of their
teacher but with cheaper convolutional blocks outperform student networks with
the original blocks and smaller architectures.

The cheap convolutional blocks we suggest are described in
Section~\ref{sec:modelcompression} as well as an overview of the methods we
employ for distillation. In Section~\ref{sec:exp} we evaluate student networks
with these blocks on CIFAR-10 and CIFAR-100~\citep{cifar}. Finally, in
Section~\ref{sec:addexp} we examine the efficacy of such networks for the tasks
of ImageNet~\citep{ILSVRC15} classification, and semantic segmentation on the
Cityscapes dataset~\citep{Cordts2016Cityscapes}. Our claims are as follows: 

\begin{itemize}
\item Greater model compression by distillation is possible by replacing
    convolutional blocks than by shrinking the architecture. 
\item Grouped convolutional blocks, with or without a bottleneck contraction,
    are an effective replacement block. 
\item This replacement is cheap in design time (substitution), and cheap in
    training complexity; it uses the same optimiser and hyperparameters as those
    used during the original training.
\end{itemize}

\section{Related Work}

The parameters in deep networks have a great deal of redundancy; it has been
shown that many of them can be predicted from a subset of
parameters~\citep{denil2013predicting}. However the challenge remains to find
good ways to exploit this redundancy without losing substantial model accuracy.
This observation, along with a desire for efficiency improvements has driven
the development of smaller, and less computationally-intensive convolutions.
One of the most prominent examples is the depthwise separable
convolution~\citep{sifre2014rigid} which applies a separate convolutional
kernel to each channel, followed by a pointwise convolution over all channels;
depthwise separable convolutions have been used in several
architectures~\citep{ioffe2015batch,chollet2016xception,xie2016aggregated}, and
were explicitly adapted to mobile devices in~\citet{howard2017mobilenets}. 

The depthwise part of this convolution is a specific case of a grouped
convolution where there are as many groups as channels; grouped convolutions
were used with a grouping of 2 (i.e.~half the channels belong to each group) in
the original AlexNet~\citep{alexnet} due to GPU memory constraints. In the work
of~\citet{deeproots17} the authors examine networks trained from scratch for
different groupings, motivated by efficiency. They found that these networks
actually generalised better than an ungrouped alternative. However, separating
the spatial and channel-wise elements is not the only way to simplify a
convolution. In \citet{jin2014flattened} the authors propose breaking up the
general 3D convolution into a set of 3 pointwise convolutions along different
axes.~\citet{wang2016factorized}~start with separable convolutions and add
topological subdivisioning, a way to treat sections of tensors separately, and
a bottleneck of the spatial dimensions. Both of these methods produce
models that are several times smaller than the original model while maintaining
accuracy.

In a separable convolution, the most expensive part is the pointwise
convolution, so it has been proposed that this operation could also be grouped
over sets of channels. However, to maintain some connections between channels,
it is helpful to add an operation mixing the channels
together~\citep{zhang2017shuffle}. More simply, a squared reduction can be
achieved by applying a bottleneck on the channels before the spatial
convolution~\citep{iandola2016squeeze,xie2016aggregated}. In this paper we
examine the potency of a separable bottleneck structure.

The work discussed thus far involves learning a compressed network from
scratch. There are alternatives to this such as retraining after reducing the
number of parameters~\citep{han2015deep,li2017sep}. We are interested in
learning our smaller network as a student through
distillation~\citep{bucila2006model,ba2013do} in conjunction with a pre-trained
teacher network.

How small can our student network be? The complex function of a large, deep
teacher network can, theoretically, be approximated by a network with a single
hidden layer with enough units~\citep{cybenko1989approximation}. The difficulty
in practice is \emph{learning that function}. Knowledge
distillation~\citep{ba2013do, hinton2015distilling} proposes to use the
information in the logits of a learnt network to train the smaller student
network. In early experiments, this was shown to be effective; networks much
smaller than the original could be trained with small increases in error.
However, modern deep architectures prove harder to compress. For example, a
deep convolutional network cannot be trivially replaced by a feedforward
architecture~\citep{urban2016do}. Two methods have been proposed to deal with
this. First, in \citet{romero2014fitnets} the authors use a linear map between
activations at intermediate points to produce an extra loss function. Second,
in~\citet{zagoruyko2016paying}, the authors choose instead to match the
activations after taking the mean over the channels and call this method~{\it
attention transfer}. In the context of this paper, we found attention transfer
to be effective in our experiments, as described in Section~\ref{sec:exp}.

\section{Compression with Cheap Convolutions}
\label{sec:modelcompression}

Given a large, deep network that performs well on a given task, we are
interested in compressing that network so that it uses fewer parameters. A
flexible and widely applicable way to reduce the number of parameters in a
model is to replace all its convolutional layers with a cheaper alternative.
Doing this replacement invariably impairs performance when this reduced network
is trained directly on the data. Fortunately, we are able to demonstrate that
modern distillation methods enable the cheaper model to have performance closer
to the original large network.

\subsection{Distillation}
\label{sub:distill}

For this paper, we utilise and compare two different distillation methods for
learning a smaller student network from a large, pre-trained teacher network:
knowledge distillation~\citep{,ba2013do,hinton2015distilling} and attention
transfer~\citep{zagoruyko2016paying}.

\paragraph{Knowledge Distillation}
Let us denote the cross entropy of two probability vectors $\pB$ and $\qB$ as
$\mathcal{L}_{CE}(\pB,\qB) = -\sum_{k} p_k \log q_k$. Assume we have a dataset
of elements, with one such element denoted $\xB$, where each element has a
corresponding one-hot class label: denote the one-hot vector corresponding to
$\xB$ by $\yB$. Given $\xB$, we have a trained teacher network $\tB =
\mathrm{teacher}(\xB)$ that outputs the corresponding logits, denoted by $\tB$;
likewise we have a student network that outputs logits  $\sB =
\mathrm{student}(\xB)$. To perform knowledge distillation we train the student
network to minimise the following loss function (averaged across all data
items):

\begin{equation}
\mathcal{L}_{KD} =  (1-\alpha)\mathcal{L}_{CE}(\yB,\sigma(\sB))\\  + 2\alpha T^{2}\mathcal{L}_{CE}\left(\sigma\left(\frac{\tB}{T}\right),\sigma\left(\frac{\sB}{T}\right)\right),
\end{equation}

where $\sigma(.)$ is the softmax function, $T$ is a temperature parameter and
$\alpha$ is a parameter controlling the ratio of the two terms. The first term
is a standard cross entropy loss penalising the student network for incorrect
classifications. The second term is minimised if the student network produces
outputs similar to that of the teacher network. The idea being that the outputs
of the teacher network contain additional, beneficial information beyond just a
class prediction.

\paragraph{Attention Transfer}

Consider some choice of layers $i =1,2,...,N_L$ in a teacher network, and the
corresponding layers in the student network. At each chosen layer $i$ of the
teacher network, collect the spatial map of the activations for channel $j$
into the vector $\aB^t_{ij}$. Let $A^t_i$ collect $\aB^t_{ij}$ for all $j$.
Likewise for the student network we correspondingly collect into $\aB^s_{ij}$
and $A_i^s$.

Now given some choice of mapping $\fB(A_i)$ that maps each collection of the
form $A_i$ into a vector, attention transfer involves learning the student
network by minimising:
\begin{equation}
\mathcal{L}_{AT} = \mathcal{L}_{CE}(\yB,\sigma(\sB))+ \beta\sum_{i=1}^{N_L}  \left\lVert \frac{\fB(A^t_{i})}{||\fB(A^t_{i})||_2} - \frac{\fB(A^s_{i})}{||\fB(A^s_{i})||_2} \right\lVert_{2}, \label{eqn:atloss}
\end{equation}
where $\beta$ is a hyperparameter. \citet{zagoruyko2016paying} recommended
using $\fB(A_i)=(1/N_{A_i})\sum_{j=1}^{N_{A_i}} \aB_{ij}^2$, where $N_{A_i}$ is
the number of channels at layer $i$. In other words, the loss targeted the
difference in the spatial map of average squared activations, where each
spatial map is normalised by the overall activation norm.

Let us examine the loss (\ref{eqn:atloss}) further. The first term is again a
standard cross entropy loss. The second term, however, ensures the spatial
distribution of the student and teacher activations are similar at selected
layers in the network, the explanation being that both networks are then
\emph{paying attention} to the same things at those layers.

\subsection{Cheap Convolutions}
\label{sec:convs}

As large fully-connected layers are no longer commonplace, convolutions make up
almost all of the parameters in modern networks.\footnote{The parameters
introduced by batch normalisation are negligible compared to those in the
convolutions. However, they are included for completeness in
Table~\ref{table:blocks}.} It is therefore desirable to make them smaller.
Here, we present several convolutional blocks that may be introduced in place
of a standard block in a network to substantially reduce its parameter cost.

\newcommand{\Nout}{N_{\mathrm{out}}}
\newcommand{\Nin}{N_{\mathrm{in}}}

First, let us consider a standard two dimensional convolutional layer that
contains $\Nout$ filters, each of size $ \Nin \times k\times k$. $\Nout$ is the
number of channels of the layer output, $\Nin$ is the number of channels of the
input, and $k\times k$ is the kernel size of each convolution. In modern
networks it is almost always the case that $\Nin\leqslant\Nout$. Let
$N=\max(\Nin,\Nout)$. Then the parameter cost of this layer is $\Nin\Nout k^2$,
and is bounded by $N^{2}k^{2}$. In a typical residual network, a block contains
two such convolutions. We will refer to this as a \emph{Standard} block $S$,
and it is outlined in Table~\ref{table:blocks}. 

\begin{figure}[!t]
          \begin{minipage}[b]{0.5\linewidth}
            \centering\includegraphics[width=0.9\textwidth]{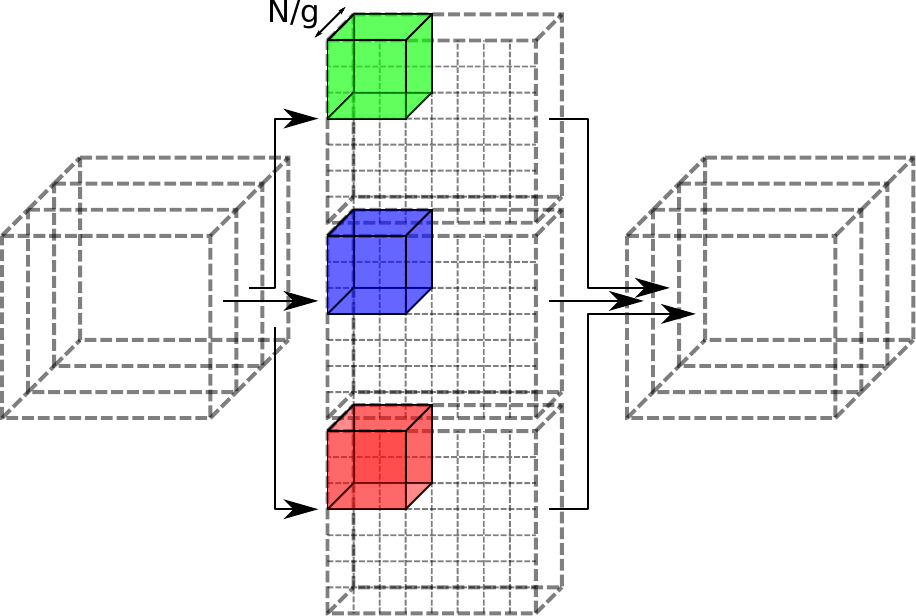}
            \subcaption{}
            \label{blocks:a}
          \end{minipage}
          \begin{minipage}[b]{0.5\linewidth}
            \centering
            \def\svgwidth{\textwidth}
            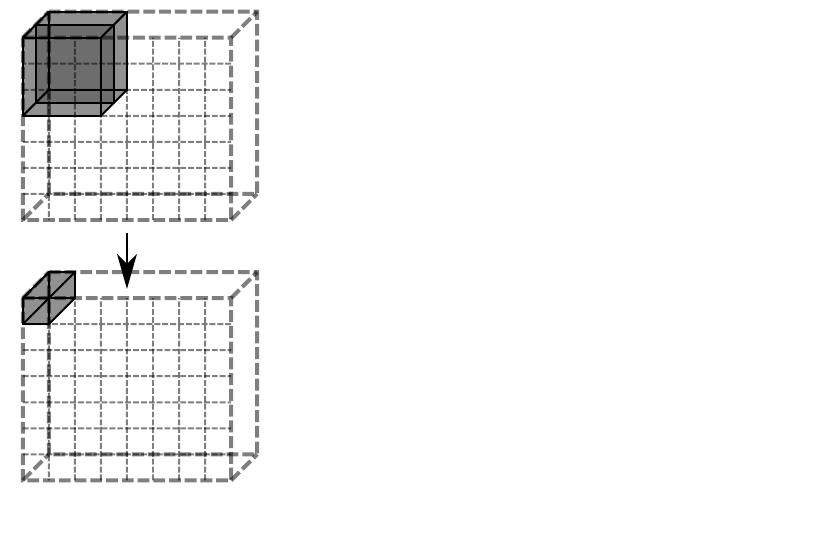
            \subcaption{}
            \label{blocks:b}
          \end{minipage}

        \caption{In ({\subref{blocks:a}}) a grouped convolution operates by
        passing independent filters over the tensor after it is separated into
        $g$ groups over the channel dimension; as each of the $g$ filters needs
        only to operate over $N/g$ channels this reduces the parameter cost of
        the layer by a factor of $g$. These can be composed into the blocks
        illustrated in ({\subref{blocks:b}}). The \emph{Grouped + Pointwise}
        ($G(g)$) block substitutes a $k \times k$ convolution with a grouped
        convolution followed by a pointwise ($1 \times 1$) convolution,
        repeating this twice. To reduce parameters further, a pointwise
        \emph{Bottleneck} can be used before the Grouped + Pointwise convolution
        ($BG(b,g)$).}
        \label{fig:blocksillustration}
\end{figure}

An alternative approach is to separate each convolution into $g$ groups, as
shown in Figure~\ref{blocks:a}. By restricting the convolutions to only mix
channels within each group, we obtain a substantial
reduction in the number of parameters for a grouped computation: for example,
for $\Nin=\Nout=N$ the cost changes from $N^{2} k^2$ for a standard layer to
$g$ groups of $(\nicefrac{N}{g})^2\,k^2$ parameter convolutions, hence reducing
the parameter cost by a factor of $g$. We can then provide some cross-group
mixing by following each grouped convolution with a pointwise convolution, with
a $N^2$ parameter cost (when $\Nin\neq\Nout$ the change in channel size occurs
across this pointwise convolution). We refer to this substitution operator as
$G(g)$ (grouped convolution with $g$ groups) and illustrate it in
Figure~\ref{blocks:b}.

In the original ResNet paper~\citep{he2016deep} the authors introduced a
bottleneck block which we have parameterised, and denoted as $B(b)$ in
Table~\ref{table:blocks}: the input first has its channels decreased by a
factor of $b$ via a pointwise convolution, before a full convolution is carried
out. Finally, another pointwise convolution brings the representation back up
to the desired $N_{out}$. We can reduce the parameter cost of this block even
further by replacing the full convolution with a grouped one; the
\emph{Bottleneck Grouped + Pointwise} block is referred to as $BG(b,g)$ and is
illustrated in Figure~\ref{blocks:b}.

These substitute blocks are compared in Table~\ref{table:blocks} and their
computational costs are given. In practice, by varying the bottleneck size and
the number of groups, network parameter numbers may vary over two orders of
magnitude; enumerated examples are given in Table~\ref{table:cifar10}.

Using grouped convolutions and bottlenecks are common methods for parameter
reduction when designing a network architecture. Both are easy to implement in
any deep learning framework. Sparsity inducing methods~\citep{han2015deep}, or
approximate layers~\citep{yang2015deep}, may also provide advantages, but these
are complementary to the approaches here. More structured reductions such as
grouped convolutions and bottlenecks can be advantageous over sparsity methods
in that the sparsity structure does not need to be stored. In the following
sections, we demonstrate that using these proposed blocks with effective model
distillation allows for substantial compression with minimal reduction in
performance.

\begin{table*}[!t]
    \caption{Convolutional Blocks used in this paper: a standard block $S$, a
        grouped + pointwise block $G$, a bottleneck block $B$, and a bottleneck
        grouped + pointwise block $BG$. Conv refers to a $k\times k$
        convolution. GConv is a grouped $k\times k$ convolution and Conv1x1 is
        a pointwise convolution. Blocks use
        pre-activations~\citep{he2016identity}: all convolutions are preceded
        by a batch-norm layer + a ReLU activation. We assume that the input and
        output to each block has $N$ channels and that channel size does not
        change over a particular convolution unless written out explicitly
        as~($x\rightarrow~y$). Where applicable, $g$ is the number of groups in
        a grouped convolution and $b$ is the bottleneck contraction. We give
        the cost of the convolutions in each block in terms of these
    parameters. The batch-norm cost at test time is also given, but is markedly smaller.}
        \vskip 0.15in
    \centering
    {\small
    \begin{tabular}{@{}lllll@{}}
        \toprule
        Block      & $S$         & $G(g)$         & $B(b)$         & $BG(b,g)$        \\ \midrule
        Structure  
                   & Conv   & GConv (g)  & Conv1x1($N\rightarrow\frac{N}{b}$)   & Conv1x1($N\rightarrow\frac{N}{b}$)   \\
                   & Conv  & Conv1x1   & Conv   & GConv(g)  \\
                   &           & GConv (g)  & Conv1x1($\frac{N}{b}\rightarrow N$)      & Conv1x1($\frac{N}{b}\rightarrow N$)      \\
                   &           & Conv1x1   &           &           \\\hline
        Conv Params & $2N^{2}k^{2}$  & $2N^{2}(\frac{k^{2}}{g}+1)$&  
        $N^{2}(\frac{k^2}{b^{2}}+\frac{2}{b})$&$N^{2}(\frac{k^2}{gb^{2}}+\frac{2}{b})$  
        \\\hline
        BN Params &$4N$&$8N$&$N(2 + \frac{4}{b})$&$N(2 + \frac{4}{b})$\\\bottomrule
    \end{tabular}
    }
    \label{table:blocks}
\end{table*}

\section{CIFAR Experiments}
\label{sec:exp}

In this section we train and evaluate a number of student networks, each
distilled from the \emph{same} large teacher network. Experiments are conducted
for both the CIFAR-10 and CIFAR-100 datasets. We distil with (i) knowledge
distillation and (ii) attention transfer. We also train the networks without
any form of distillation (i.e.\ from scratch) to observe whether the
distillation process is necessary to obtain good performance. In this way we
demonstrate that the high performance comes from the distillation, and cannot
be achieved by directly training the student networks using the data.

For comparison we also study student networks with smaller architectures (i.e.\
fewer layers/filters) than the teacher. This enables us to test if the block
transformations we propose are key, or it is simply a matter of distilling
networks with smaller numbers of parameters. We compare the smaller student
architectures with student architectures implementing cheap, substitute
convolutional blocks, but with the same architecture as the teacher. The
different convolutional blocks are summarised in Table~\ref{table:blocks} and
the student networks are described in detail in Section~\ref{sec:nets}. Results
are given in Table~\ref{table:cifar10} and Figure~\ref{fig:cifar10plot}. These
results are discussed in detail in Section~\ref{sec:analysis}.

\subsection{Network Descriptions}
\label{sec:nets}

For our experiments we utilise the Wide Residual Network (WRN)
architecture~\citep{zagoruyko2016wide}; the bulk of the network lies in
its~\{$conv2$, $conv3$, $conv4$\} groups and the network depth $d$ determines
the number of convolutional blocks $n$ in these groups as $n = (d-4)/6$. The
network width, denoted by $k$, affects the channel size of the filters in these
blocks. Note that when we employ attention transfer the student and teacher
outputs of groups \{$conv2$, $conv3$, $conv4$\} are used as \{$A_{1}$, $A_{2}$,
$A_{3}$\}~in the second term of Equation (\ref{eqn:atloss}) with $N_L = 3$.

\noindent For our teacher network we use WRN-40-2 (a WRN with depth 40 and
width multiplier 2) with standard ($S$) blocks. $3 \times 3$ kernels are used for all
non-pointwise convolutions in our  student and teacher networks unless stated
otherwise. For our student networks we use:
\begin{itemize}
  \item WRN-40-1, 16-2, and 16-1 with $S$ blocks. These are student networks
      that are thinner and/or more shallow than the teacher and represent
        typical student networks used.
  \item WRN-40-2 with $S$ blocks where the $3\times3$ kernels have been replaced with $2\times2$ dilated kernels (as described in~\citet{dilated2016}). This allows us to see if it possible to naively reduce parameters by effectively zeroing out elements of standard kernel.
  \item WRN-40-2 using a bottleneck block $B$ with $2\times$ and $4\times$ channel contraction~($b$).
  \item WRN-40-2 using a grouped + pointwise block $G$ for group sizes ($g$) \{2, 4, 8, 16, N/16, N/8, N/4, N/2, N\} where N is the number of channels in a given block. This allows us to explore the spectrum between full convolutions ($g = 1$) and fully separable convolutions ($g = N$).
  \item WRN-40-2 with a bottleneck grouped + pointwise block $BG$. We use $b=2$ with groups sizes of \{2, 4, 8, 16, M/16, M/8, M/4, M/2, M\} where $M = N/b$ is the number of channels~{\it after the bottleneck}. We use this notation so that $g = M$ represents fully separable convolutions and we can easily denote divisions thereof. $BG(4,M)$ is also used to observe the effect of extreme compression.
\end{itemize}

\paragraph{Implementation Details}
For training we used minibatches of size 128. Before each minibatch, the images
were padded by $4\times4$ zeros, and then a random  $32\times32$ crop was
taken. Each image was left-right flipped with a probability of a half. Networks
were trained for 200 epochs using SGD with momentum fixed at 0.9 with an
initial learning rate of 0.1. The learning rate was reduced by a factor of 0.2
at the start of epochs 60, 120, and 160. For knowledge distillation we set
$\alpha$ to 0.9 and used a temperature of 4. For attention transfer $\beta$ was
set to 1000. The code to reproduce these experiments is available at 
{\url{https://github.com/BayesWatch/pytorch-moonshine}}.

\subsection{Analysis and Observations}
\label{sec:analysis}

\begin{figure*}
   \begin{tabular}{c}
   \includegraphics[width=\textwidth]{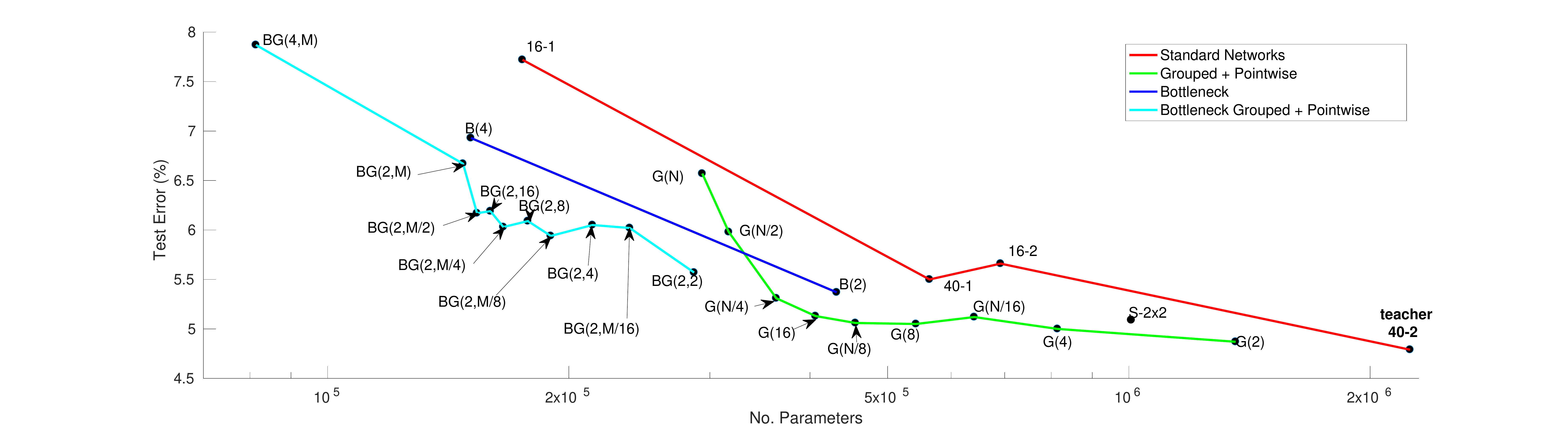}\\
   (a)\\
    \includegraphics[width=\textwidth]{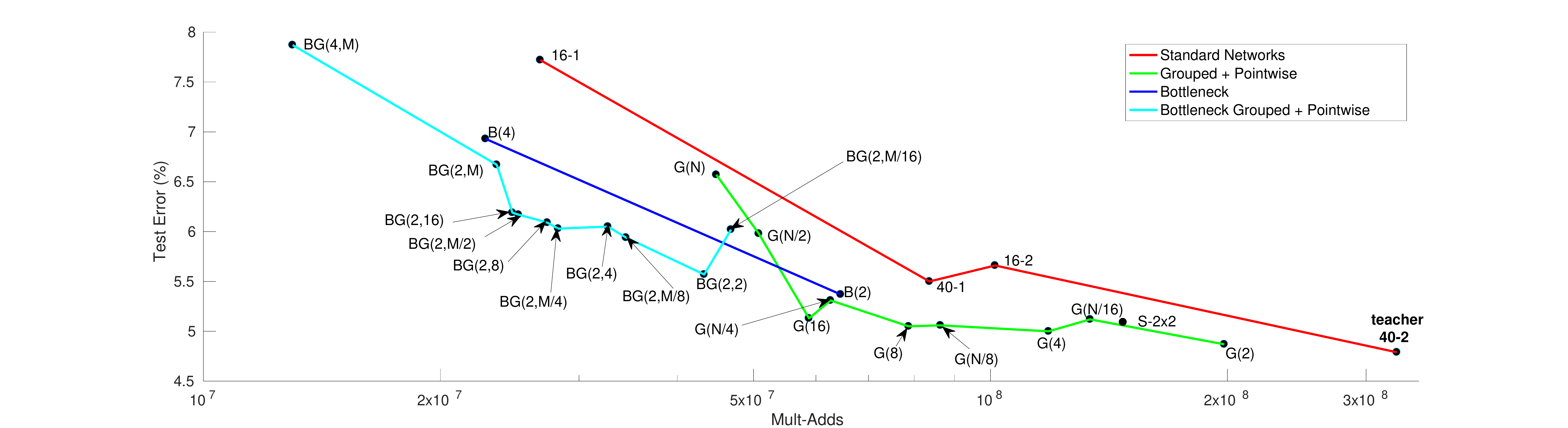}\\
   (b)\\
	\end{tabular}
  \caption{Test Error vs. (a) No.~parameters and (b) Mult-adds for student networks learnt with
    attention transfer on CIFAR-10. Note that the x-axes are log-scaled. Points on the red curve correspond to networks with $S$
    convolutional blocks and reduced architectures. All other networks have the
    same WRN-40-2 architecture as the teacher but with cheap convolutional
    blocks: $G$ (green), $B$ (blue), and $BG$ (cyan). The blocks are described in Table~\ref{table:blocks}. Notice that the student networks with
    cheap blocks outperform those with smaller architectures and standard
    convolutions for a given parameter budget or mult-add budget.}
  \label{fig:cifar10plot}
\end{figure*}

Figure~\ref{fig:cifar10plot}a compares the parameter cost of each student
network (on a log scale) against the test error on CIFAR-10 obtained with
attention transfer. On this plot, the ideal network would lie in the
bottom-left corner (few parameters, low error). What is fascinating is that
almost every network with the same architecture as the teacher, but with cheap
convolutional blocks (those on the blue, green, and cyan lines) performs better
for a given parameter budget than the reduced architecture networks with
standard blocks (the red line). $BG(2,2)$ outperforms $16$-$2$ (5.57\% vs.
5.66\%) despite having considerably fewer parameters (287K vs. 692K). Several
of the networks with $BG$ blocks both significantly outperform $16$-$1$ and use
fewer parameters. 

It is encouraging that significant compression is possible with only small
losses; several networks perform almost as well as the teacher with
considerably fewer parameters -- $G(N/8)$ has an error of $5.06\%$, close to
that of the teacher, but has just over a fifth of the parameters. $BG(2,M/8)$
has less than a tenth of the parameters of the teacher, for a cost of $1.15\%$
increase in error. Even
simply switching all convolutions with smaller, dilated equivalents
($S-2\times2$) allows one to use half the parameters for a similar performance.

An important lesson can be learnt regarding grouped + pointwise convolutions.
They are often used in their depthwise-separable~\citep{chollet2016xception}
form ($g = N$). However, the networks with half, or quarter that number of
groups perform substantially better for a modest increase in
parameters.~$G(N/4)$ has 363K parameters compared to the 294K of $G(N)$ but has
an error that is~$1.26\%$ lower. The number of groups is an easy parameter to
tune to trade some performance for a smaller network. Grouped + pointwise
convolutions also work well in conjunction with a bottleneck of size 2,
although for large bottlenecks the error increases significantly, as can be
seen for $BG(4,M)$. Despite this, it is still of comparable performance to
$16$-$1$ with half the parameters. Similar trends are observed for CIFAR-100 in
Table~\ref{table:cifar10}b.

We also observe that training a student with attention transfer is
substantially better than using knowledge distillation, or simply training from
scratch. Consider Table~\ref{table:cifar10}, which shows the attention transfer
errors of Figure~\ref{fig:cifar10plot} (the AT column) alongside those of
networks trained with knowledge distillation (KD), and no distillation i.e.\ from scratch (Scr) for CIFAR-10 and CIFAR-100. In all cases, the student network trained with
attention transfer is better than the student network trained by itself -- the
distillation process appears to be necessary. Some performances are
particularly impressive; on CIFAR-10, for $G(2)$ blocks the error is only
$0.08\%$ higher than the teacher despite the network having 60\% of the
parameters.

These results support our claim that greater model compression through
distillation is possible by substituting the convolutional blocks in a network,
rather than by shrinking its architecture. We have also demonstrated that the
blocks outlined in Table~\ref{table:blocks} are suitable substitutes. By
observing Figure~\ref{fig:cifar10plot}b we can also see that our networks
with cheap, substitute blocks utilise fewer mult-add operations than their
standard equivalents, which roughly corresponds to a faster runtime. However,
it is worth noting that actual runtime on a given platform or device is
dependent on specifics (memory paging, choice of libraries etc.), so mult-adds
are not always fully indicative of runtime, but are a decent approximation in a
platform/implementation-agnostic setting.

\begin{table*}[!t]
    \caption{Student Network test error on CIFAR-10/100. Each network is a WideResNet with its depth-width (D-W) given in the first column, and with its block
        type (corresponding to Table~\ref{table:blocks} in the second. $N$ refers to the channel width of each
        block, and $M$ refers to the channel width after the bottleneck where
        applicable. The total parameter cost of the networks for CIFAR-10 is given, as well as the number of mult-add operations they use. Note that CIFAR-100 networks use an extra 11.6K parameters and mult-adds over their CIFAR-10 equivalents as they have a larger linear classification layer.
        Errors are reported for (i) learning with no distillation i.e.\ from scratch (Scr),
        (ii) knowledge distillation with a teacher (KD), and attention
        transfer with a teacher (AT). The same teacher is used for training,
        and is given in the first row. This table shows that (i) through attention
        transfer it is possible to cut the number of parameters of a network, but
        retain high performance and (ii) for a similar number of parameters,
        students with cheap convolutional blocks outperform those with expensive
        convolutions and smaller architectures.}
        \vskip 0.05in
    \centering

    \begin{tabular}{@{}ll|rrrrr|rrr@{}}
    
    \multicolumn{2}{c}{}&\multicolumn{5}{c}{CIFAR-10}  &\multicolumn{3}{c}{CIFAR-100} \\
        \toprule
D-W    & Block      & Params (K) & MAdds (M) & Scr & KD   & AT   & Scr & KD    & AT    \\ \midrule
        {\bf T} 40-2   & S          & 2243.5     & 328.3     & 4.79    & --   & --   & 23.85   & --    & --    \\
        \hline
16-2 & S          & 691.7  & 101.4 & 6.53 & 6.03 & 5.66 & 27.63 & 27.97 & 27.24 \\
40-1 & S          & 563.9  & 83.6  & 6.48 & 6.39 & 5.50  & 29.64 & 30.21 & 28.24 \\
16-1 & S          & 175.1    & 26.8  & 8.81 & 8.75 & 7.72 & 34.00    & 37.28 & 33.74 \\
\hline
40-2 & S-2x2      & 1007.1   & 147.4 & 5.89 & 6.03 & 5.09 & 27.20  & 26.98 & 26.09 \\
\hline
40-2 & G(2)       & 1359.0 & 198.1 & 5.30  & 5.37 & 4.87 & 25.94 & 24.92 & 24.45 \\
40-2 & G(4)       & 814.7  & 118.5 & 5.50  & 5.81 & 5.00    & 26.20  & 25.48 & 25.30  \\
40-2 & G(8)       & 542.5  & 78.7  & 5.92 & 5.72 & 5.05 & 26.49 & 26.64 & 25.71 \\
40-2 & G(16)      & 406.4  & 58.8  & 6.65 & 6.38 & 5.13 & 28.85 & 27.10  & 26.34 \\
40-2 & G(N/16)    & 641.3  & 133.9 & 5.72 & 5.72 & 5.12 & 27.08 & 26.11 & 25.78 \\
40-2 & G(N/8)     & 455.8  & 86.4  & 6.07 & 5.61 & 5.06 & 27.85 & 27.05 & 26.15 \\
40-2 & G(N/4)     & 363.1    & 62.6  & 6.93 & 6.45 & 5.31 & 28.91 & 27.93 & 26.85 \\
40-2 & G(N/2)     & 316.7  & 50.8  & 7.12 & 6.83 & 5.98 & 30.24 & 28.89 & 28.54 \\
40-2 & G(N)       & 293.5  & 44.8  & 8.51 & 8.01 & 6.57 & 31.84 & 29.99 & 30.06 \\
\hline
40-2 & B(2)       & 431.8  & 64.5  & 6.36 & 6.28 & 5.37 & 28.27 & 28.08 & 26.68 \\
40-2 & B(4)       & 150.9  & 22.8  & 7.94 & 7.83 & 6.93 & 31.63 & 33.63 & 30.56 \\
\hline
40-2 & BG(2,2)    & 286.7  & 43.3  & 6.12 & 6.25 & 5.57 & 28.51 & 28.82 & 28.28 \\
40-2 & BG(2,4)    & 214.1  & 32.7  & 6.75 & 6.75 & 6.05 & 29.39 & 29.25 & 28.54 \\
40-2 & BG(2,8)    & 177.8  & 27.3  & 6.94 & 6.98 & 6.09 & 30.21 & 29.34 & 28.89 \\
40-2 & BG(2,16)   & 159.7  & 24.7  & 6.77 & 6.97 & 6.19 & 30.57 & 30.54 & 29.46 \\
40-2 & BG(2,M/16) & 238.3  & 46.8  & 6.26 & 6.50  & 6.02 & 29.69 & 28.69 & 29.05 \\
40-2 & BG(2,M/8)  & 189.9  & 34.4  & 6.75 & 6.49 & 5.94 & 29.09 & 29.13 & 28.16 \\
40-2 & BG(2,M/4)  & 165.7  & 28.2  & 7.06 & 7.15 & 6.03 & 30.42 & 30.28 & 28.60  \\
40-2 & BG(2,M/2)  & 153.6  & 25.1  & 7.45 & 7.47 & 6.17 & 30.44 & 30.66 & 29.51 \\
40-2 & BG(2,M)    & 147.6  & 23.6  & 7.95 & 7.99 & 6.67 & 30.90  & 31.18 & 30.03 \\
40-2 & BG(4,M)    & 81.4   & 13.0  & 9.04 & 8.61 & 7.87 & 33.64 & 37.34 & 32.89\\
\bottomrule
    \end{tabular}
    \label{table:cifar10}
\end{table*}

\section{Additional Experiments}
\label{sec:addexp}

Section~\ref{sec:exp} demonstrates the effectiveness of \emph{cheapening}
convolutions for CIFAR classification. In this section, we apply this method to
two further problems. Firstly, in Section~\ref{sec:imagenet} we examine whether
the benefits observed hold for large-scale image classification on
ImageNet~\citep{ILSVRC15} where there are far more classes (1000), and the
images are significantly larger. Secondly, in Section~\ref{sec:seg} we cheapen
the convolutions of a network trained for semantic segmentation.

\subsection{ImageNet}
\label{sec:imagenet}

Our experiments use a pre-trained ResNet-34~\citep{he2016deep} (21.8M
parameters) as a teacher and we train several networks using attention transfer
(AT). We compare student networks that have the architecture of ResNet-34, with
\emph{cheaper} convolutions, to those that have reduced architectures, and full
convolutions. Note, that the bulk of the parameters in a ResNet are contained
in four groups, as opposed to the three of a WideResNet. We train the
following student networks: (i) ResNet-18, (ii) ResNet-18 with the channel
widths of the last three groups halved (Res18-0.5), ResNet-34 with each
convolutional block replaced by (iii) a $G(N)$ block\footnote{As the
convolutional blocks in the teacher do not use pre-activations, the $G$ blocks
used here are modified accordingly (BN + ReLU now come after each convolution).
This also applies to the networks in Section~\ref{sec:seg}.} and (iv) a $G(4)$
block. Validation errors for these networks are available in
Table~\ref{table:imagenet}.

Consider Res34-G(N) and Res18-0.5, which both have roughly the same parameter cost
($\sim$3M). After distillation, the former has a significantly lower top-5 error
(10.66\% vs. 15.02\%). This again supports our claim that is is preferable to
cheapen convolutions, rather than shrink the network architecture. Res34-G(N)
trained from scratch has a noticeably higher top-5 error (12.26\%), it benefits
from distillation. Conversely, distillation makes Res18-0.5 slightly worse,
suggesting that it has no further representational capacity. 

Res34-G(4) similarly outperforms Res18 (these are roughly similar in cost at
8.1M and 11.7M parameters respectively), although in this case the latter does
benefit from distillation. It is intriguing that Res34-G(4) trained from
scratch is actually on par with the original teacher (having a 0.12\% lower
top-1 error, and a 0.05\% higher top-5 error) despite having 13 million fewer
parameters; this generalisation capability of grouped convolutions in networks
has been observed previously by~\citet{deeproots17}. Distillation is able to
push its performance slightly further to the point that its top-5 error
surpasses that of the teacher (8.43\% vs. 8.57\%).

\paragraph{Implementation Details}

Models were trained for 100 epochs using SGD with an initial learning rate of
0.1, momentum of 0.9 and weight decay of $10^{-4}$. The learning rate was
reduced by a factor of 10 every 30 epochs. Minibatches of size 256 were used
across 4 GPUs. When trained with a teacher, an additional AT loss was used with
the outputs of the four groups of each ResNet. $\beta$ was set to 750 so that
the total contribution of the AT loss was the same as in Section~\ref{sec:exp}.

\begin{table}
    \caption{Top 1 and Top 5 classification errors (\%) on the validation set
    of ImageNet for models (i) trained from scratch, and (ii) those trained
with attention transfer with ResNet-34 (Res34) as a teacher. Res18 refers to a
ResNet-18, and Res18-0.5 is a ResNet-18 where the channel width in the last
three groups is halved. Res34-G($x$) is a ResNet-34 with each convolutional
block replaced by a G($x$) block. We can observe that for a particular
parameter budget (3M or $\sim$10M) the networks with cheap replacement blocks
outperform those with reduced architectures. These trends follow for mult-adds. Note that the Res34 and Res18
scratch results were obtained from pre-trained PyTorch models.}
        \vskip 0.05in
    \centering
    \begin{tabular}{@{}lrrrrrr@{}}
          \multicolumn{3}{c}{}&\multicolumn{2}{c}{Scratch} &\multicolumn{2}{c}{AT} \\
        \toprule
        Model & Params & Mult-Adds & Top 1& Top 5& Top 1&Top 5\\ \midrule
        Res34~{\bf T}&21.8M&3.669G&26.73&8.57&--&--\\
        \hline
        Res18&11.7M&1.818G&30.36&11.02&29.18&10.05\\
         Res34-G(4)&8.1M&1.395G&26.61&8.62&26.58&8.43\\
         \hline
         Res18-0.5&3.2M&909M&36.96&15.01&37.20&15.02\\
        Res34-G(N)&3.1M&559M&32.98&12.26&30.16&10.66\\

        \bottomrule

    \end{tabular}

    \label{table:imagenet}
\end{table}
\subsection{Semantic Segmentation}
\label{sec:seg}

We have shown that cheapening the convolutions of a network, coupled with a
good distillation process, has allowed for a substantial reduction in the
number of network parameters in return for a small drop in performance.
However, the networks trained thus far have all had the same task -- image
classification. Here, we take an existing network, trained for the task of
semantic segmentation and apply our method to distil it.

For our teacher network we use an ERFNet~\citep{Romera2017a,Romera2017b} that
has been trained \emph{from scratch} on the Cityscapes
dataset~\citep{Cordts2016Cityscapes} -- a collection of images of urban street
scenes, in which each pixel has been labelled as one of 19 classes. The bulk of
an ERFNet is made up of standard residual blocks where each full convolution
has been replaced by a pair of 1D alternatives: a $3\times1$ convolution
followed by a $1\times3$ convolution. The second such pair in each block is
often dilated. To \emph{cheapen} this network for use as a student, we replace
each block with a $G(N)$ block, maintaining the dilations where appropriate.

We use the same optimiser, and training schedule as for the original ERFNet.
When training the student the only difference is the addition of an attention
transfer term (see Equation~\ref{eqn:atloss}) between several of the feature
maps in the final loss. The models are evaluated using class
Intersection-over-Union (IoU) accuracy on the validation set, and the results
can be found in Table~\ref{table:cityscapes}. 

In~\citet{Romera2017b} the authors detail how ERFNet is designed with
efficiency in mind. With only one training run and no tuning, we are able to
reduce the number of parameters to one quarter of the original for
a modest drop in performance.

\begin{table}
    \caption{IoU accuracy (\%) on the validation set of Cityscapes for (i) ERFNet, and (ii) ERFNet with replacement blocks (ERFNet-G(N)). For ERFNet-G(N) the accuracy when trained from scratch (Scratch IoU), and when used as a student with the original ERFNet as a teacher (AT IoU) is given.}
    \vskip 0.15in
    \centering
    \begin{tabular}{@{}lrrrr@{}}
        \toprule
        Model       & Params   & Mult-Adds  & Scratch IoU & AT IoU\\ \midrule
        ERFNet      & 2.06M    &3.73G      & 70.59       & -- \\
        ERFNet-G(N) & 0.49M    & 1.19G       & 65.29       & 68.11\\
        \bottomrule
    \end{tabular}

    \label{table:cityscapes}
\end{table}
\vspace{-2mm}

\paragraph{Implementation Details}
Models were trained using the same optimiser, schedule, and image scaling and
augmentation as in the ERFNet paper~\citep{Romera2017b} with the attention
transfer loss for the case of ERFNet-G(N) as a student. For encoder training,
the outputs of layers 7, 12, and 16 were used for attention transfer with
$\beta = 1000$. For decoder training, the outputs of layers 19 and 22 were also
used and $\beta$ was dropped to 600 (so that the contribution of this term
remains this same).
\vspace{-3mm}

\section{Conclusion}
\vspace{-3mm}

After training a large, deep model it may be prohibitively time consuming to
design a model compression strategy in order to deploy it. On many problems, it
may also be more difficult to achieve the desired performance with a smaller
model. We have demonstrated a model compression strategy that is fast to apply,
and doesn't require any additional engineering for both image classification
and semantic segmentation. Furthermore, the optimisation algorithm of the
larger model is sufficient to train the cheaper student model.

The cheap convolutions used in this paper were chosen for their ease of
implementation. Future work could investigate more complicated approximate
operations, such as those described in \citet{moczulski2015acdc}; which could
make a difference for the $1 \times 1$ convolutions in the final layers of a
network. One could also make use of custom blocks generated through a large
scale black box optimisation as in~\citet{zoph2017learning}. Equally, there are
many methods for low rank approximations that could be
applicable~\citep{sainath2013low,jaderberg2014speeding,garipov2016ultimate}. We
hope that this work encourages others to consider \emph{cheapening their
convolutions} as a compression strategy.

\paragraph{Acknowledgements.}
This project has received funding from the European Union’s Horizon 2020
research and innovation programme under grant agreement No. 732204 (Bonseyes).
This work is supported by the Swiss State Secretariat for Education‚ Research
and Innovation (SERI) under contract number 16.0159. The opinions expressed and
arguments employed herein do not necessarily reflect the official views of
these funding bodies. The authors are grateful to Sam Albanie, Luke Darlow, Jack Turner, and the anonymous reviewers for their helpful suggestions.

\FloatBarrier

\bibliography{mybib}

\begin{thebibliography}{37}
\providecommand{\natexlab}[1]{#1}
\providecommand{\url}[1]{\texttt{#1}}
\expandafter\ifx\csname urlstyle\endcsname\relax
  \providecommand{\doi}[1]{doi: #1}\else
  \providecommand{\doi}{doi: \begingroup \urlstyle{rm}\Url}\fi

\bibitem[Ba \& Caruana(2014)Ba and Caruana]{ba2013do}
Ba, L.~J. and Caruana, R.
\newblock Do deep nets really need to be deep?
\newblock In \emph{Advances in Neural Information Processing Systems}, 2014.

\bibitem[Buciluǎ et~al.(2006)Buciluǎ, Caruana, and
  Niculescu-Mizil]{bucila2006model}
Buciluǎ, C., Caruana, R., and Niculescu-Mizil, A.
\newblock Model compression.
\newblock In \emph{ACM SIGKDD {I}nternational {C}onference on {K}nowledge
  {D}iscovery and {D}ata {M}ining}, 2006.

\bibitem[Chollet(2016)]{chollet2016xception}
Chollet, F.
\newblock {Xception}: Deep learning with depthwise separable convolutions.
\newblock \emph{arXiv:1610.02357}, 2016.

\bibitem[Cordts et~al.(2016)Cordts, Omran, Ramos, Rehfeld, Enzweiler, Benenson,
  Franke, Roth, and Schiele]{Cordts2016Cityscapes}
Cordts, M., Omran, M., Ramos, S., Rehfeld, T., Enzweiler, M., Benenson, R.,
  Franke, U., Roth, S., and Schiele, B.
\newblock The cityscapes dataset for semantic urban scene understanding.
\newblock In \emph{Proceedings of the IEEE Conference on Computer Vision and
  Pattern Recognition}, 2016.

\bibitem[Cybenko(1989)]{cybenko1989approximation}
Cybenko, G.
\newblock Approximation by superpositions of a sigmoidal function.
\newblock \emph{Mathematics of Control, Signals, and Systems (MCSS)},
  2\penalty0 (4):\penalty0 303--314, 1989.

\bibitem[Denil et~al.(2013)Denil, Shakibi, Dinh, Ranzato, and
  de~Freitas]{denil2013predicting}
Denil, M., Shakibi, B., Dinh, L., Ranzato, M., and de~Freitas, N.
\newblock Predicting parameters in deep learning.
\newblock In \emph{Advances in Neural Information Processing Systems}, 2013.

\bibitem[Garipov et~al.(2016)Garipov, Podoprikhin, Novikov, and
  Vetrov]{garipov2016ultimate}
Garipov, T., Podoprikhin, D., Novikov, A., and Vetrov, D.~P.
\newblock Ultimate tensorization: compressing convolutional and {FC} layers
  alike.
\newblock \emph{arXiv:1611.03214}, 2016.

\bibitem[Han et~al.(2016)Han, Mao, and Dally]{han2015deep}
Han, S., Mao, H., and Dally, W.~J.
\newblock Deep compression: Compressing deep neural networks with pruning,
  trained quantization and {H}uffman coding.
\newblock In \emph{International Conference on Learning Representations}, 2016.

\bibitem[He et~al.(2016{\natexlab{a}})He, Zhang, Ren, and Sun]{he2016deep}
He, K., Zhang, X., Ren, S., and Sun, J.
\newblock Deep residual learning for image recognition.
\newblock In \emph{Proceedings of the IEEE Conference on Computer Vision and
  Pattern Recognition}, 2016{\natexlab{a}}.

\bibitem[He et~al.(2016{\natexlab{b}})He, Zhang, Ren, and Sun]{he2016identity}
He, K., Zhang, X., Ren, S., and Sun, J.
\newblock Identity mappings in deep residual networks.
\newblock In \emph{European Conference on Computer Vision}, 2016{\natexlab{b}}.

\bibitem[Hinton et~al.(2015)Hinton, Vinyals, and Dean]{hinton2015distilling}
Hinton, G., Vinyals, O., and Dean, J.
\newblock Distilling the knowledge in a neural network.
\newblock \emph{arXiv:1503.02531}, 2015.

\bibitem[Howard et~al.(2017)Howard, Zhu, Chen, Kalenichenko, Wang, Weyand,
  Andreetto, and Adam]{howard2017mobilenets}
Howard, A.~G., Zhu, M., Chen, B., Kalenichenko, D., Wang, W., Weyand, T.,
  Andreetto, M., and Adam, H.
\newblock Mobile{N}ets: Efficient convolutional neural networks for mobile
  vision applications.
\newblock \emph{arXiv:1704.04861}, 2017.

\bibitem[Iandola et~al.(2016)Iandola, Moskewicz, Ashraf, Han, Dally, and
  Keutzer]{iandola2016squeeze}
Iandola, F.~N., Moskewicz, M.~W., Ashraf, K., Han, S., Dally, W.~J., and
  Keutzer, K.
\newblock Squeeze{N}et: Alex{N}et-level accuracy with $50\times$ fewer
  parameters and $<1${M}{B} model size.
\newblock \emph{arXiv:1602.07360}, 2016.

\bibitem[Ioannou et~al.(2017)Ioannou, Robertson, Cipolla, and
  Criminisi]{deeproots17}
Ioannou, Y., Robertson, D., Cipolla, R., and Criminisi, A.
\newblock Deep roots: Improving {CNN} efficiency with hierarchical filter
  groups.
\newblock In \emph{Proceedings of the IEEE Conference on Computer Vision and
  Pattern Recognition}, 2017.

\bibitem[Ioffe \& Szegedy(2015)Ioffe and Szegedy]{ioffe2015batch}
Ioffe, S. and Szegedy, C.
\newblock Batch normalization: Accelerating deep network training by reducing
  internal covariate shift.
\newblock In \emph{International Conference on Machine Learning}, 2015.

\bibitem[Jaderberg et~al.(2014)Jaderberg, Vedaldi, and
  Zisserman]{jaderberg2014speeding}
Jaderberg, M., Vedaldi, A., and Zisserman, A.
\newblock Speeding up convolutional neural networks with low rank expansions.
\newblock In \emph{British Machine Vision Conference}, 2014.

\bibitem[Jin et~al.(2015)Jin, Dundar, and Culurciello]{jin2014flattened}
Jin, J., Dundar, A., and Culurciello, E.
\newblock Flattened convolutional neural networks for feedforward acceleration.
\newblock In \emph{International Conference on Learning Representations}, 2015.

\bibitem[Krizhevsky(2009)]{cifar}
Krizhevsky, A.
\newblock Learning multiple layers of features from tiny images.
\newblock Master's thesis, University of Toronto, 2009.

\bibitem[Krizhevsky et~al.(2012)Krizhevsky, Sutskever, and Hinton]{alexnet}
Krizhevsky, A., Sutskever, I., and Hinton, G.
\newblock Image{N}et classification with deep convolutional neural networks.
\newblock In \emph{Advances in Neural Information Processing Systems}, 2012.

\bibitem[LeCun et~al.(2015)LeCun, Bengio, and Hinton]{lecun2015deep}
LeCun, Y., Bengio, Y., and Hinton, G.
\newblock Deep learning.
\newblock \emph{Nature}, 521\penalty0 (7553):\penalty0 436--444, 2015.

\bibitem[Li et~al.(2017)Li, Wang, Lv, and Yang]{li2017sep}
Li, Z., Wang, X., Lv, X., and Yang, T.
\newblock {S}{E}{P}-{N}ets: Small and effective pattern networks.
\newblock \emph{arXiv:1706.03912}, 2017.

\bibitem[Moczulski et~al.(2016)Moczulski, Denil, Appleyard, and
  de~Freitas]{moczulski2015acdc}
Moczulski, M., Denil, M., Appleyard, J., and de~Freitas, N.
\newblock {ACDC:} a structured efficient linear layer.
\newblock In \emph{International Conference on Learning Representations}, 2016.

\bibitem[Romera et~al.(2017{\natexlab{a}})Romera, {\'A}lvarez, Bergasa, and
  Arroyo]{Romera2017a}
Romera, E., {\'A}lvarez, J.~M., Bergasa, L.~M., and Arroyo, R.
\newblock Efficient convnet for real-time semantic segmentation.
\newblock \emph{IEEE Intelligent Vehicles Symposium}, 2017{\natexlab{a}}.

\bibitem[Romera et~al.(2017{\natexlab{b}})Romera, {\'A}lvarez, Bergasa, and
  Arroyo]{Romera2017b}
Romera, E., {\'A}lvarez, J.~M., Bergasa, L.~M., and Arroyo, R.
\newblock {E}{R}{F}{N}et: Efficient residual factorized convnet for real-time
  semantic segmentation.
\newblock \emph{IEEE Transactions on Intelligent Transportation Systems},
  2017{\natexlab{b}}.

\bibitem[Romero et~al.(2015)Romero, Ballas, Kahou, Chassang, Gatta, and
  Bengio]{romero2014fitnets}
Romero, A., Ballas, N., Kahou, S.~E., Chassang, A., Gatta, C., and Bengio, Y.
\newblock Fit{N}ets: Hints for thin deep nets.
\newblock In \emph{International Conference on Learning Representations}, 2015.

\bibitem[Russakovsky et~al.(2015)Russakovsky, Deng, Su, Krause, Satheesh, Ma,
  Huang, Karpathy, Khosla, Bernstein, Berg, and Fei-Fei]{ILSVRC15}
Russakovsky, O., Deng, J., Su, H., Krause, J., Satheesh, S., Ma, S., Huang, Z.,
  Karpathy, A., Khosla, A., Bernstein, M., Berg, A.~C., and Fei-Fei, L.
\newblock {ImageNet Large Scale Visual Recognition Challenge}.
\newblock \emph{International Journal of Computer Vision (IJCV)}, 115\penalty0
  (3):\penalty0 211--252, 2015.
\newblock \doi{10.1007/s11263-015-0816-y}.

\bibitem[Sainath et~al.(2013)Sainath, Kingsbury, Sindhwani, Arisoy, and
  Ramabhadran]{sainath2013low}
Sainath, T.~N., Kingsbury, B., Sindhwani, V., Arisoy, E., and Ramabhadran, B.
\newblock Low-rank matrix factorization for deep neural network training with
  high-dimensional output targets.
\newblock In \emph{IEEE International Conference on Acoustics, Speech and
  Signal Processing}, 2013.

\bibitem[Sifre(2014)]{sifre2014rigid}
Sifre, L.
\newblock \emph{Rigid-Motion Scattering for Image Classification}.
\newblock PhD thesis, \'Ecole Polytechnique, 2014.

\bibitem[Urban et~al.(2017)Urban, Geras, Kahou, Aslan, Wang, Caruana, Mohamed,
  Philipose, and Richardson]{urban2016do}
Urban, G., Geras, K.~J., Kahou, S.~E., Aslan, O., Wang, S., Caruana, R.,
  Mohamed, A., Philipose, M., and Richardson, M.
\newblock Do deep convolutional nets really need to be deep and convolutional?
\newblock In \emph{International Conference on Learning Representations}, 2017.

\bibitem[Wang et~al.(2016)Wang, Liu, and Foroosh]{wang2016factorized}
Wang, M., Liu, B., and Foroosh, H.
\newblock Design of efficient convolutional layers using single intra-channel
  convolution, topological subdivisioning and spatial bottleneck structure.
\newblock \emph{arXiv:1608.04337}, 2016.

\bibitem[Xie et~al.(2017)Xie, Girshick, Doll{\'{a}}r, Tu, and
  He]{xie2016aggregated}
Xie, S., Girshick, R., Doll{\'{a}}r, P., Tu, Z., and He, K.
\newblock Aggregated residual transformations for deep neural networks.
\newblock In \emph{Proceedings of the IEEE Conference on Computer Vision and
  Pattern Recognition}, 2017.

\bibitem[Yang et~al.(2015)Yang, Moczulski, Denil, de~Freitas, Smola, Song, and
  Wang]{yang2015deep}
Yang, Z., Moczulski, M., Denil, M., de~Freitas, N., Smola, A., Song, L., and
  Wang, Z.
\newblock Deep fried convnets.
\newblock In \emph{Proceedings of the IEEE International Conference on Computer
  Vision}, 2015.

\bibitem[Yu \& Koltun(2016)Yu and Koltun]{dilated2016}
Yu, F. and Koltun, V.
\newblock Multi-scale context aggregation by dilated convolutions.
\newblock In \emph{International Conference on Learning Representations}, 2016.

\bibitem[Zagoruyko \& Komodakis(2016)Zagoruyko and
  Komodakis]{zagoruyko2016wide}
Zagoruyko, S. and Komodakis, N.
\newblock Wide residual networks.
\newblock In \emph{British Machine Vision Conference}, 2016.

\bibitem[Zagoruyko \& Komodakis(2017)Zagoruyko and
  Komodakis]{zagoruyko2016paying}
Zagoruyko, S. and Komodakis, N.
\newblock Paying more attention to attention: Improving the performance of
  convolutional neural networks via attention transfer.
\newblock In \emph{International Conference on Learning Representations}, 2017.

\bibitem[{Zhang} et~al.(2018){Zhang}, {Zhou}, {Lin}, and
  {Sun}]{zhang2017shuffle}
{Zhang}, X., {Zhou}, X., {Lin}, M., and {Sun}, J.
\newblock Shuffle{N}et: An extremely efficient convolutional neural network for
  mobile devices.
\newblock In \emph{Proceedings of the IEEE Conference on Computer Vision and
  Pattern Recognition}, 2018.

\bibitem[Zoph et~al.(2018)Zoph, Vasudevan, Shlens, and Le]{zoph2017learning}
Zoph, B., Vasudevan, V., Shlens, J., and Le, Q.~V.
\newblock Learning transferable architectures for scalable image recognition.
\newblock In \emph{Proceedings of the IEEE Conference on Computer Vision and
  Pattern Recognition}, 2018.

\end{thebibliography}
\bibliographystyle{icml2018}

\end{document}